\documentclass[letterpaper, 10 pt, conference]{ieeeconf}  
\usepackage{graphicx} 
\usepackage{stfloats}
\usepackage{amsmath} 
\usepackage{algorithm,algpseudocode}
\usepackage{wrapfig}
\usepackage{amssymb}
\usepackage{graphicx}
\usepackage{wrapfig}
\usepackage{cite}
\usepackage{booktabs}
\usepackage{pifont}
\usepackage{multirow}
\usepackage{booktabs}  
\usepackage{array}
\usepackage{makecell}
\usepackage{pifont}

\IEEEoverridecommandlockouts                              

\title{\LARGE \bf
VoxAfford: Multi-Scale Voxel-Token Fusion for Open-Vocabulary 3D Affordance Detection
}

\author{Haowen~Sun, Shaolong~Zhang, Mingyang~Li, Chengzhong~Ma, Xinzhe~Chen, Qiongjie~Cui,\\ Xingyu~Chen, Zeyang~Liu, Xuguang~Lan$^{*}$
\thanks{The authors are with the National Key Laboratory of Human-Machine Hybrid Augmented Intelligence, Institute of Artificial Intelligence and Robotics, Xi'an Jiaotong University, Xi'an, 710049}}

\begin{document}

\maketitle
\thispagestyle{empty}
\pagestyle{empty}

\begin{abstract}
Open-vocabulary 3D affordance detection requires localizing interaction regions on point clouds given novel affordance descriptions. Recent methods extend multimodal large language models (MLLMs) with special output tokens that are decoded into segmentation masks. However, these tokens are produced through autoregressive generation, which models sequential dependencies rather than spatial neighborhood relations, leaving them semantically rich but spatially impoverished for 3D localization. We propose \textbf{Vox}el-enhanced \textbf{Afford}ance detection (\textbf{VoxAfford}), which bypasses this bottleneck by injecting multi-scale geometric features from a frozen pre-trained 3D VQVAE encoder into the output tokens after generation. Each output token uses its affordance semantics as a query to retrieve relevant geometric patterns from its paired voxel scale via cross-attention, with a learned compatibility gate controlling the injection strength. The enhanced tokens are then aggregated into a spatially-aware affordance prompt through semantic-conditioned attention and propagated alongside per-point features to generate the final mask. Experiments on open-vocabulary affordance detection tasks show that VoxAfford achieves state-of-the-art performance with approximately an 8\% improvement in mIoU, and real robot experiments confirm zero-shot transfer to novel objects.
\end{abstract}

\section{Introduction}

Affordance detection identifies regions on objects that support specific interactions~\cite{rana2024learning, wu2025afforddp, xu2025a0, fan2025video}. Unlike semantic segmentation, where labels are primarily determined by object identity, affordance labels arise from the joint consideration of task intent, object properties, and local geometric configurations. This task-conditioned geometric reasoning makes affordance fundamentally harder than category-level labeling. 3D point clouds preserve the geometric structure needed for this reasoning and are therefore widely adopted for affordance detection~\cite{tang2025uad, qian2024affordancellm, wei20253daffordsplat}. Building on this representation, most existing methods train per-point classifiers on a fixed set of affordance categories~\cite{deng20213d, mo2022o2o, zhu2025grounding}. These methods cannot generalize to affordance types absent from the training set. Open-vocabulary affordance detection removes this restriction by accepting free-form language queries, but it also introduces a new challenge: the model must determine what geometric conditions a novel affordance requires and locate the corresponding regions on the object.

\begin{figure}
    \centering
    \includegraphics[width=0.45\textwidth]{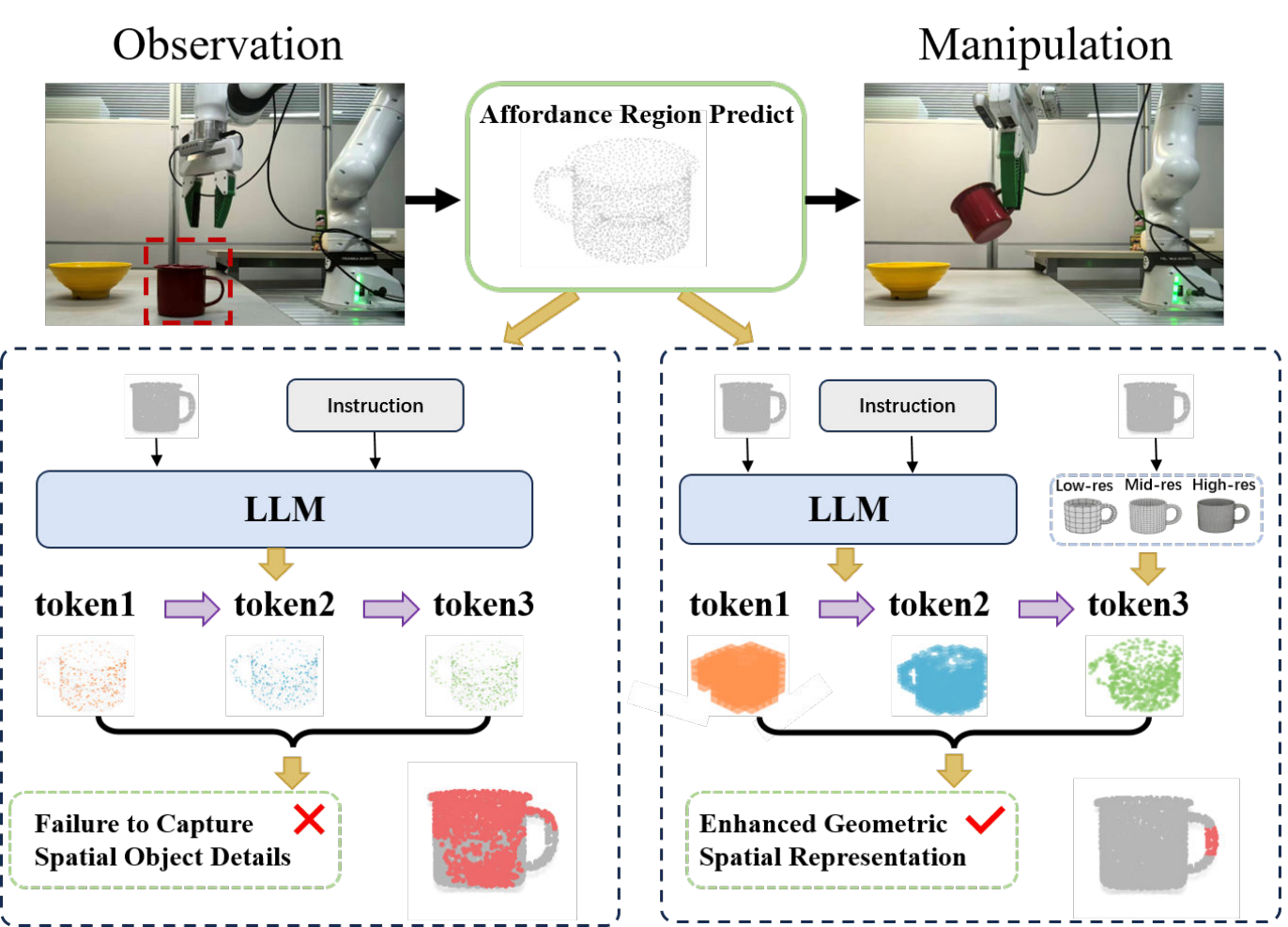}
    \caption{\textbf{MLLM-based affordance region prediction for robotic manipulation}. Given a scene observation, to execute a pouring instruction the robot first needs to predict the affordance region (e.g., the handle of a mug for grasping) on the target object. Left: Existing methods rely solely on MLLM output tokens to decode affordance masks, but the autoregressive bottleneck limits the geometric information carried by these tokens, leading to inaccurate predictions. Right: VoxAfford fuses hierarchical voxel features with the MLLM output tokens, producing spatially accurate affordance masks.
    }
    \label{fig1}
\end{figure}

Recent methods adopt multimodal large language models (MLLMs) to enable open-vocabulary affordance detection~\cite{wang2026affordance, he2026task, van2024open, wu2025open}, but the output tokens that guide mask prediction carry limited geometric information about the target region. In these methods, the LLM output vocabulary is extended with special tokens whose hidden representations are decoded into segmentation masks by a downstream decoder~\cite{yu2025seqafford, chu20253d, zhang2025openhoi}. As shown in Figure~\ref{fig1}, the tokens capture the semantic meaning of the queried affordance, but they encode little about where the corresponding region is located in 3D space. Some methods enrich the 3D input to the LLM with stronger encoders so that the model receives more geometric detail~\cite{chen2024recon3d, zhou2023uni3d, mar2015multi, zhang2025positional}. Others apply auxiliary spatial losses on the output tokens to encourage them to encode positional information~\cite{huang2025reason3d,lai2024lisa}. However, both approaches require geometric information to pass through the autoregressive generation process before reaching the output tokens. The output tokens therefore remain spatially impoverished regardless of how much geometric information is supplied at the input or how the output is supervised.

We attribute this limitation to a structural mismatch between autoregressive generation and 3D geometry. Autoregressive models build representations along a sequential token order, whereas 3D geometric properties are determined by spatial neighborhood relations. The language embedding space therefore lacks an inductive bias for encoding spatial topology. With sufficient data, autoregressive models can in principle acquire spatial structure, as shown in large-scale image generation~\cite{yu2022scaling, ding2021cogview, gafni2022make, esser2021taming}. However, 3D affordance datasets are orders of magnitude smaller, making it impractical to learn spatial representations through fine-tuning alone. A more practical alternative is to supply pre-learned geometric representations from an external source. This shifts the learning problem from acquiring geometric representations to aligning them with affordance semantics. Such alignment is especially critical in open-vocabulary settings, where the decoder must generalize to novel affordance types and cannot rely on memorized associations from training.

In this paper, we propose \textbf{Vox}el-enhanced \textbf{Afford}ance detection (\textbf{VoxAfford}), which injects hierarchical voxel features into the LLM output tokens after generation to supply the missing geometric information. A frozen pre-trained 3D Vector Quantized Variational Autoencoder (VQVAE) encoder~\cite{chen2025sar3d, qi2024shapellm, ye2025shapellm} extracts geometric features at three resolutions ($16^3$, $32^3$, $64^3$). These resolutions encode complementary categories of geometric information, ranging from global occupancy topology to local surface properties. Each output token serves as a semantic query that retrieves relevant geometric patterns from its paired voxel scale through cross-attention. A learned gate then measures the compatibility between the affordance semantics and the retrieved geometry, controlling how much geometric content is injected into each token. The enhanced tokens are then aggregated and propagated to a per-point decoder for affordance prediction. Experiments on open-vocabulary affordance detection tasks show that VoxAfford achieves state-of-the-art performance with approximately an 8\% improvement in mIoU. Real robot experiments confirm model transfer to novel objects under both full and partial point cloud observations.

Our main contributions are as follows.
\begin{itemize}
    \item We propose VoxAfford, which injects hierarchical voxel features from a frozen pre-trained VQVAE encoder into the LLM output tokens after generation through semantics-conditioned cross-attention. A learned compatibility gate controls the injection strength based on the alignment between affordance semantics and retrieved geometry. This supplies structured geometric information to the tokens without modifying the LLM architecture.
    \item We pair each output token with a designated voxel resolution, enabling each token to specialize at a distinct spatial granularity.
    \item VoxAfford achieves state-of-the-art performance on open-vocabulary affordance detection tasks. Real robot experiments confirm model transfer to novel objects under both full and partial point cloud observations.
\end{itemize}

\begin{figure*}
    \centering
    \includegraphics[width=0.95\textwidth]{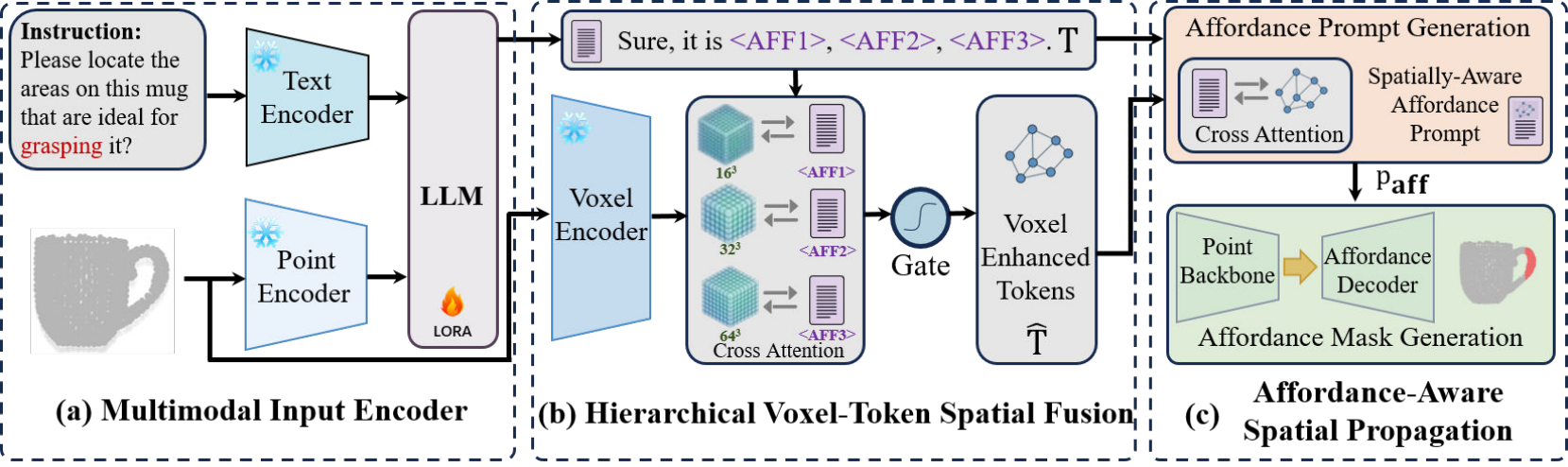}
    \caption{\textbf{Overview of the VoxAfford framework}. (a) Multimodal Input Encoder. A Text Encoder and a Point Encoder encode the affordance query and object point cloud, respectively, and feed them into an LLM fine-tuned with LoRA that autoregressively generates affordance tokens. (b) Hierarchical Voxel-Token Spatial Fusion. A Voxel Encoder extracts multi-scale geometric features from the voxelized point cloud; cross attention fuses these voxel features with the LLM affordance tokens to produce Voxel Enhanced Tokens. (c) Affordance-Aware Spatial Propagation. Cross attention between the Voxel Enhanced Tokens and point features generates a Spatially-Aware Affordance Prompt, which is combined with a Point Backbone to drive an Affordance Decoder for final Affordance Mask Generation.
    }
    \label{fig2}
\end{figure*}

\section{Related Work}

\subsection{3D Affordance Detection}

Affordance, first introduced by Gibson~\cite{gibson1966senses}, describes the interaction possibilities that objects offer to agents. Early work focuses on 2D images, predicting pixel-level affordance regions from RGB observations~\cite{tang2025affordgrasp, zhu2025afford, do2018affordancenet}. Extending this to 3D, Deng et al.~\cite{deng20213d} construct the 3D AffordanceNet benchmark and predict per-point affordance heatmaps on object point clouds. Subsequent methods improve performance through diverse strategies: analyzing actionable parts for 3D manipulation~\cite{mo2021where2act, wu2021vat}, modeling inter-category affordance relations via graph reasoning~\cite{yang2023grounding}, aligning language and point cloud representations~\cite{li2024laso}, and addressing partial-view generalization~\cite{tang2025uad, wu2025afforddp}. However, these methods operate under closed-set settings with fixed label taxonomies and cannot generalize to novel affordance types absent from training.

To enable open-vocabulary generalization, while early attempts map 3D features to pre-trained 2D VLM spaces~\cite{peng2023openscene, shao2025great, van2024open}, recent methods progressively reformulate affordance detection as instruction-conditioned reasoning with 3D multimodal LLMs~\cite{hong20233d, xu2024pointllm}. Specifically, 3D-AffordanceLLM~\cite{qian2024affordancellm} feeds point cloud features and natural language queries into an MLLM, which predicts affordance masks through a dedicated output token. SeqAfford~\cite{yu2025seqafford} further introduces sequential output tokens to capture multi-region affordances. These methods successfully leverage the language reasoning capability of LLMs for zero-shot generalization~\cite{chen2024ll3da, chen2024grounded}, but the output tokens that guide mask prediction are produced entirely within the language embedding space and carry limited 3D spatial information. Our work addresses this spatial deficiency by injecting geometric representations into the output tokens after generation, bypassing the autoregressive bottleneck rather than relying on richer inputs or auxiliary output losses.

\subsection{Multimodal LLMs for 3D Understanding}

Multimodal LLMs have achieved strong results in 2D vision-language tasks~\cite{li2023blip, liu2023visual, zhu2023minigpt, bai2023qwen} and have been progressively extended to 3D understanding. PointLLM~\cite{xu2024pointllm} and 3D-LLM~\cite{hong20233d} enable 3D question answering and captioning by projecting point cloud features into the LLM embedding space. For 3D dense prediction, a central challenge is how to equip the model with sufficient spatial understanding to produce accurate per-point outputs.

On the input side, prior works explore stronger 3D encoders to provide richer geometric features to the LLM. Point-based backbones such as PointNet++~\cite{qi2017pointnet++} and PointBERT~\cite{yu2022point} extract local geometric features from raw coordinates, while Uni3D~\cite{zhou2023uni3d} and Recon++~\cite{chen2024recon3d} learn transferable 3D representations through large-scale pre-training. Beyond point-level encoding, ShapeLLM~\cite{qi2024shapellm} feeds hierarchical voxel features from a pre-trained 3D VQVAE encoder~\cite{chen2025sar3d} into the LLM as input tokens, demonstrating that structured voxel representations improve 3D-language alignment. However, in all these methods the geometric information must still pass through the autoregressive generation process before reaching the output.
On the output side, LISA~\cite{lai2024lisa} introduces a special token whose hidden representation is decoded into a segmentation mask, and Reason3D~\cite{huang2025reason3d} extends this paradigm to 3D with auxiliary spatial losses to encourage positional encoding in the output tokens. Despite these efforts on both sides, the output tokens are inherently produced within the language embedding space and lack explicit 3D spatial structure. Unlike ShapeLLM~\cite{qi2024shapellm}, which uses voxel features as LLM input, our method injects multi-scale voxel features into the output tokens after generation through cross-attention, addressing the spatial limitation at the representation level without requiring geometric information to traverse the autoregressive bottleneck.

\section{Method}

Given an object point cloud $P_c \in \mathbb{R}^{N \times 3}$ and a natural language affordance query $Q_a$, VoxAfford predicts a per-point binary mask $M_a \in \{0,1\}^N$ indicating the queried affordance region. Figure~\ref{fig2} illustrates the overall architecture. VoxAfford processes the input through two parallel encoding paths. On the language-semantic path, a text encoder and a point encoder $f_{\text{pe}}$ encode the affordance query and the object point cloud, respectively, and feed them into an LLM fine-tuned with LoRA, which autoregressively generates $K$ affordance tokens. These are special tokens appended to the LLM vocabulary whose hidden representations encode the model's semantic understanding of the queried affordance and will guide the downstream mask prediction. On the geometric path, the same point cloud is voxelized and fed to a frozen pre-trained voxel VQVAE encoder, which extracts multi-scale geometric features at three resolutions. The voxel encoder is pre-trained on large-scale shape reconstruction, and its hierarchical downsampling path naturally provides geometric representations ranging from global shape topology to local surface detail.

The two paths converge in the hierarchical voxel-token spatial fusion module (Section~\ref{sec:fusion}), where cross-attention fuses the voxel features with the LLM affordance tokens, with a learned compatibility gate controlling the injection strength, producing voxel-enhanced tokens. This supplies structured 3D spatial information directly to the affordance tokens without requiring it to pass through the autoregressive generation process. The voxel-enhanced tokens, together with dense point features extracted by a point backbone $f_{\text{PB}}$, are then processed by the affordance-aware spatial propagation module (Section~\ref{sec:propagation}), which generates a spatially-aware affordance prompt through cross-attention between the voxel-enhanced tokens and point features, and combines it with the point backbone to drive an affordance decoder $f_{\text{AFD}}$ for final mask prediction. Section~\ref{sec:training} describes the two-stage training procedure.

\subsection{Hierarchical Voxel-Token Spatial Fusion}
\label{sec:fusion}

The goal of voxel-token fusion is to let each affordance token selectively retrieve geometric patterns from the voxel features based on its affordance semantics, so that different affordance queries attend to different spatial regions on the same object. The voxel features encode 3D geometry in a structured grid that preserves spatial neighborhood relations absent in unordered point clouds, but they are agnostic to the queried affordance. The fusion therefore combines the token's semantic specificity with the voxel features' spatial structure through multi-scale cross-attention retrieval, followed by selective injection via a learned compatibility gate.

\paragraph{Multi-scale voxel feature extraction and retrieval.}
Affordance prediction requires reasoning over geometric relations at multiple granularities. For example, localizing a graspable region relies more on local surface detail, while identifying a sittable surface depends more on global shape understanding. A single voxel resolution cannot provide both. We extract features at three resolutions ($16^3$, $32^3$, $64^3$) from the hierarchical downsampling path of a frozen pre-trained 3D VQVAE encoder. The large-scale shape reconstruction pre-training provides geometric priors that transfer to novel objects. These features capture geometric information at different spatial granularities.

Each affordance token is paired with one voxel resolution and retrieves geometric patterns from its paired scale through cross-attention:
\begin{equation}
    \mathbf{g}_j = \text{CrossAttention}(\mathbf{t}_j, \, \phi_j(\mathbf{V}_j), \, \phi_j(\mathbf{V}_j)), \quad j \in \{1,2,3\}
\end{equation}
where $\mathbf{t}_j$ is the $j$-th LLM affordance token, $\mathbf{V}_j$ denotes the voxel features at the $j$-th resolution, and $\phi_j$ is a per-scale linear projection. Different affordance queries produce different attention distributions over the same voxel grid, retrieving geometric patterns relevant to the queried interaction.

\paragraph{Selective geometric injection.}
The retrieved geometric patterns are not all relevant to the queried affordance. Injecting irrelevant geometry would introduce noise into the token representation. We therefore gate the injection based on the compatibility between the affordance semantics and the retrieved geometry. A scalar gate $\alpha_j$ is computed from both the semantic token and the retrieved geometry:
\begin{equation}
    \alpha_j = \sigma\!\left(\mathbf{w}_j^\top \left[f_s(\mathbf{t}_j);\, f_g(\mathbf{g}_j)\right] + b_j\right)
\end{equation}
\begin{equation}
    \hat{\mathbf{t}}_j = \mathbf{t}_j + \alpha_j \cdot \mathbf{W}_j \mathbf{g}_j,
\end{equation}
where $\sigma$ denotes the sigmoid function, $f_s$ and $f_g$ are linear projections and $\mathbf{w}_j$, $b_j$, $\mathbf{W}_j$ are per-scale learnable parameters. The gate opens when the retrieved geometry is compatible with the affordance semantics and closes when it is irrelevant. We initialize the gate near zero following ControlNet~\cite{zhang2023adding}, so that the injection starts with no effect and increases gradually during training.

The fixed pairing between each token and its voxel resolution constrains the category of geometry available to each token, providing an inductive bias that drives multi-scale specialization without explicit supervision. We denote the resulting representations $\hat{\mathbf{T}} = \{\hat{\mathbf{t}}_1, \hat{\mathbf{t}}_2, \hat{\mathbf{t}}_3\}$ as the voxel-enhanced tokens.

\subsection{Affordance-Aware Spatial Propagation}
\label{sec:propagation}

The voxel-token fusion module produces voxel-enhanced tokens $\hat{\mathbf{T}} = \{\hat{\mathbf{t}}_1, \hat{\mathbf{t}}_2, \hat{\mathbf{t}}_3\}$ that carry both affordance semantics and multi-scale geometric information. The point backbone $f_{\text{PB}}$ independently extracts dense point embeddings $\mathbf{F}$ from the input point cloud. The spatial propagation module bridges these two representations to produce the per-point affordance mask.

We observe that the voxel-enhanced tokens and the original LLM tokens $\mathbf{T}$ are suited for different roles in this process. The enhanced tokens $\hat{\mathbf{T}}$ encode spatial awareness of where the affordance region is likely located through their injected geometric structure, making them effective for generating a global spatial prompt that guides the decoder toward the correct region. The original tokens $\mathbf{T}$ preserve the LLM's semantic understanding of what the queried affordance means, making them better suited for conditioning individual point features with affordance semantics. If the roles were reversed, the prompt would lack spatial information for localization, while the injected geometric signals would conflict with the point backbone's own spatial encoding. We therefore design a dual-pathway propagation that routes each token type to its appropriate role.

\paragraph{Affordance prompt generation.}
Different affordance types rely on different spatial scales, so the aggregation of voxel-enhanced tokens should be conditioned on the affordance semantics rather than using a fixed weighting scheme. We compute an affordance query from the mean of the original LLM tokens and attend over the voxel-enhanced tokens to produce the affordance prompt:
\begin{equation}
    \mathbf{p}_{\text{aff}} = \text{LN}\!\left(\mathbf{q}_{\text{aff}} + \text{CrossAttention}(f_q(\frac{1}{K}\sum_{j=1}^{K} \mathbf{t}_j), \, \hat{\mathbf{T}}, \, \hat{\mathbf{T}})\right),
\end{equation}
where $K$ is the number of affordance tokens, $f_q$ is a learnable projection, and LN denotes layer normalization. Through this attention mechanism, a grasp query naturally assigns higher weight to the $64^3$ token for fine-grained surface detail, while a sit query attends more to the $16^3$ token for global shape context.

\paragraph{Affordance mask generation.}
The affordance prompt provides global spatial guidance but does not modify individual point representations. To condition each point with affordance semantics, we inject the original LLM tokens $\mathbf{T}$ into the dense point features $\mathbf{F}$ through cross-attention with a zero-initialized residual connection:
\begin{equation}
    \hat{\mathbf{F}} = \mathbf{F} + f_{\text{zero}}(\text{CrossAttention}(\mathbf{F}, \, \mathbf{T}, \, \mathbf{T})),
\end{equation}
where $f_{\text{zero}}$ is a linear projection initialized with zero weights, following the same principle as the compatibility gate in Section~\ref{sec:fusion}. The conditioned point features and the affordance prompt are then decoded into the per-point affordance mask:
\begin{equation}
    M_a = f_{\text{AFD}}(\mathbf{p}_{\text{aff}}, \hat{\mathbf{F}}),
\end{equation}
where $f_{\text{AFD}}$ follows the two-way transformer architecture~\cite{kirillov2023segment}.

\subsection{Training Strategy}
\label{sec:training}

Our goal is to equip the model with the capability of predicting per-point affordance masks for natural language queries, including affordance types absent from the training set. Achieving this capability requires both basic segmentation ability and affordance-specific spatial reasoning. We therefore adopt a two-stage training strategy. The first stage pre-trains the point backbone and decoder on a part segmentation task to establish segmentation ability. The second stage trains the full model on affordance data, where the voxel-token fusion and spatial propagation modules learn to inject and propagate geometric information for affordance prediction.

\paragraph{Segmentation pre-training.}
We pre-train the point backbone $f_{\text{PB}}$ and the affordance decoder $f_{\text{AFD}}$ on the PartNet dataset~\cite{mo2019partnet}, which provides part-level annotations for 3D objects. Given a text description of an object part and the corresponding point cloud, the model predicts a binary segmentation mask of the described part. The point encoder, VQVAE encoder, and LLM backbone are frozen during this stage. This pre-training stage is necessary because affordance training data is limited in scale, making it difficult for the point backbone and decoder to simultaneously learn basic segmentation ability and affordance-specific spatial reasoning from scratch. Pre-training on PartNet first establishes robust segmentation capability, allowing the subsequent affordance training stage to focus on learning the alignment between affordance semantics and geometric structure. The pre-trained weights of $f_{\text{PB}}$ and $f_{\text{AFD}}$ initialize the corresponding modules in the next stage.

\paragraph{Affordance training with voxel-token fusion.}
The full model is trained on affordance data with the point encoder, VQVAE encoder, and LLM backbone frozen. The LLM is adapted with LoRA~\cite{hu2022lora}. The trainable components include the voxel-token fusion module, the spatial propagation module, $f_{\text{PB}}$, and $f_{\text{AFD}}$. The training objective combines the language modeling loss with mask prediction losses:
\begin{equation}
    \mathcal{L} = \lambda_1 \mathcal{L}_{\text{CE}} + \lambda_2 \mathcal{L}_{\text{BCE}} + \lambda_3 \mathcal{L}_{\text{Dice}},
\end{equation}
where $\mathcal{L}_{\text{CE}}$ is the auto-regressive cross-entropy loss for text generation, $\mathcal{L}_{\text{BCE}}$ is the per-point binary cross-entropy loss, and $\mathcal{L}_{\text{Dice}}$ is the dice loss for mask prediction.

\begin{figure*}
    \centering
    \includegraphics[width=0.9\textwidth]{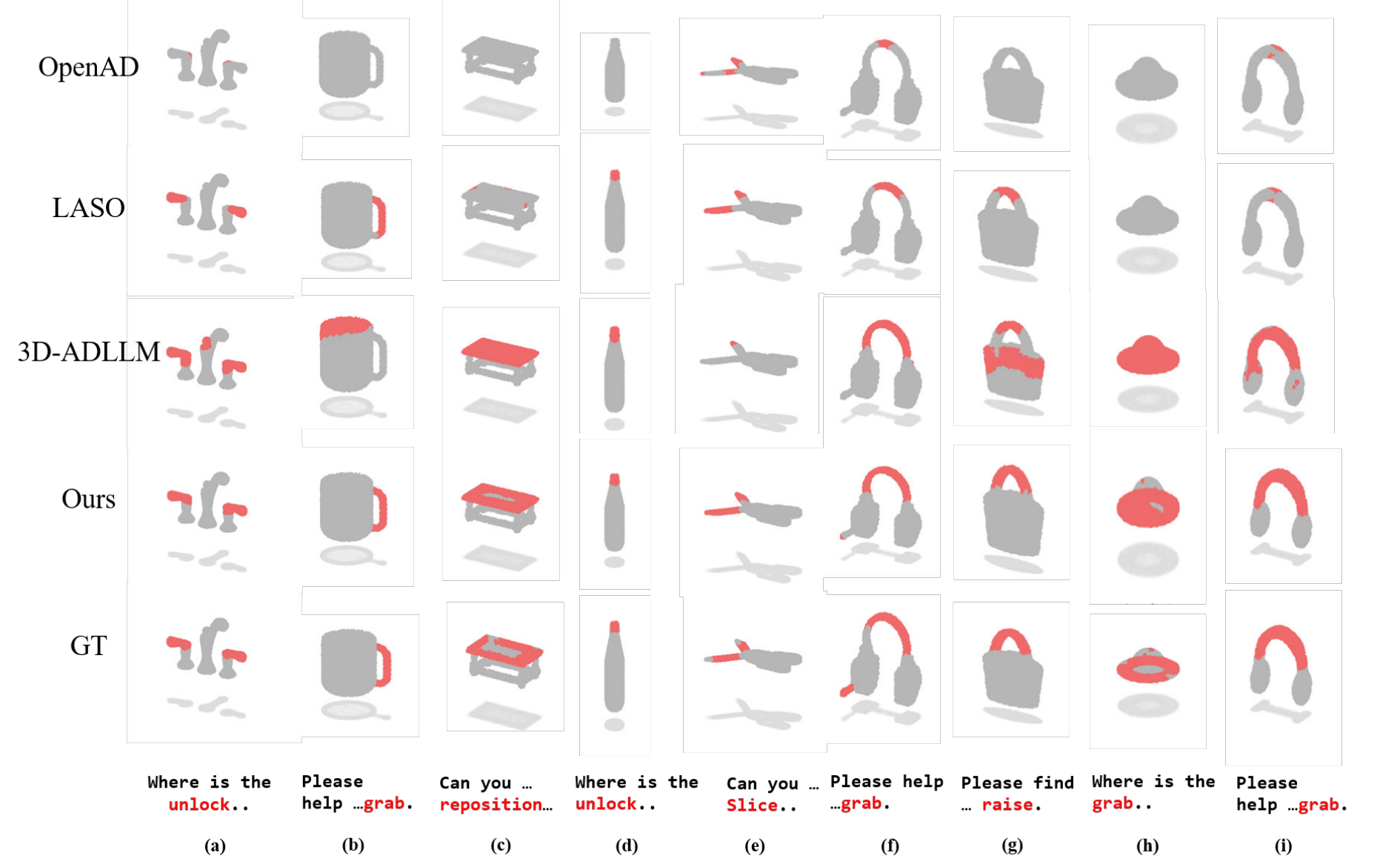}
    \caption{The visualization results of our VoxAfford compared with others.
    }
    \label{fig3}
\end{figure*}


\section{Experiments}

We design experiments to answer three questions: (1) Can VoxAfford achieve open-vocabulary zero-shot affordance prediction and transfer to real robot manipulation on novel objects? (2) Do the three tokens capture geometric relations at distinct spatial granularities? (3) Does the spatial deficiency originate from the autoregressive generation process rather than insufficient input encoding, and are the component designs justified?

\subsection{Experimental Setup}

\paragraph{Datasets.}
We use two types of training data. For segmentation pre-training, we use PartNet~\cite{mo2019partnet}, which contains 573,585 part instances across 25,571 3D models and 24 object categories. For affordance training, we use 42,119 question-answer pairs from the 3D AffordanceNet dataset~\cite{deng20213d}, covering 23 object classes and 36 affordance types. We evaluate on the OpenAD benchmark following the open-set and close-set splits defined in~\cite{deng20213d}. In the open-set evaluation, the test affordance categories are disjoint from the training set, meaning the model has never seen examples of these affordance types during training.

\paragraph{Evaluation metrics.}
We report both class-level and instance-level metrics. For class-level evaluation, we use mIoU$^c$ (mean IoU over all classes), Acc$^c$ (overall accuracy), and mAcc$^c$ (mean accuracy over all classes). For instance-level evaluation, we use mIoU$^i$, mAcc$^i$, mPrec$^i$ (mean precision), mRec$^i$ (mean recall), and mAP$_{50}^i$ (mean average precision at 50\% IoU). Following~\cite{chu20253d}, we exclude the ``none'' category from metric computation.

\paragraph{Baselines.}
We compare with recent methods for zero-shot 3D affordance detection: ZSLPC~\cite{cheraghian2019zero}, TZSLPC~\cite{cheraghian2020transductive}, 3DGenZ~\cite{michele2021generative}, OpenAD (PointNet++ and DGCNN variants)~\cite{nguyen2023open}, IAGNet~\cite{yang2023grounding}, LASO~\cite{li2024laso}, ShapeLLM~\cite{qi2024shapellm}, and 3D-ADLLM~\cite{chu20253d}. All baselines are evaluated under the same open-set zero-shot protocol.

\paragraph{Implementation details.}
We use Phi-4-mini-instruct~\cite{abdin2024phi} as our base LLM. For the point encoder, we adopt PointBERT~\cite{yu2022point}, pre-trained with ULIP-2 \cite{xue2024ulip} in the ModelNet dataset \cite{wu20153d}. Additionally, we utilize the Point Transformer \cite{zhao2021point} as the backbone for our point segmentation model. The LLM is fine-tuned with LoRA (rank 32, alpha 48). The VQVAE encoder is loaded from the ShapeLLM-Omni~\cite{ye2025shapellm} pre-trained checkpoint and remains frozen throughout training.

\subsection{Comparison Results}

Table~\ref{tab:main_class} and Table~\ref{tab:main_instance} report class-level and instance-level results on the OpenAD open-set zero-shot benchmark. VoxAfford outperforms all baselines under both full-view and partial-view settings. At the class level, VoxAfford improves mIoU$^c$ over 3D-ADLLM by 4.05 on full-view and 4.62 on partial-view. At the instance level, mIoU$^i$ improves by 8.84 and mAP$_{50}^i$ improves by 10.46 on full-view. The gains concentrate on localization-sensitive metrics such as mIoU and mAP$_{50}$, while per-point accuracy remains comparable across methods. This pattern is consistent with the design of our voxel-token fusion module. The injected voxel features provide explicit 3D spatial structure that helps the model recover more complete affordance regions, whereas per-point classification already benefits from the LLM's semantic understanding and receives less additional benefit from geometric injection.
Under partial-view, VoxAfford outperforms all baselines on mIoU and mAP$_{50}$ under both views, while maintaining competitive accuracy. The VQVAE encoder extracts hierarchical features from the voxelized input regardless of view completeness. The resulting geometric representation preserves global shape topology even when the point cloud is incomplete, which allows the fusion module to supply useful spatial cues under partial observation.

\begin{table}[t]
\caption{Zero-shot open-vocabulary detection results over all classes. The overall results of all comparative methods, the best results are in \textbf{bold}. $^*$ denotes results without fine-tuning.}
\label{tab:main_class}
\centering
\small
\setlength{\tabcolsep}{1pt}
\begin{tabular}{l|ccc|ccc}
\hline
\multirow{2}{*}{Method} & \multicolumn{3}{c|}{Full-view} & \multicolumn{3}{c}{Partial-view} \\
 & mIoU$^c$ & Acc$^c$ & mAcc$^c$ & mIoU$^c$ & Acc$^c$ & mAcc$^c$ \\
\hline
TZSLPC & 3.86 & -- & 10.37 & 4.14 & -- & 8.49 \\
3DGenZ & 6.46 & -- & 18.33 & 6.03 & -- & 15.86 \\
ZSLPC & 9.97 & -- & 18.70 & 9.52 & -- & 17.16 \\
\hline
ShapeLLM$^*$ & 0.88 & 0.28 & 0.99 & 1.49 & 1.35 & 1.70 \\
OpenAD-PointNet++ & 13.53 & 3.97 & 16.40 & 11.29 & 2.41 & 13.88 \\
OpenAD-DGCNN & 11.15 & 3.84 & 13.86 & 8.04 & 1.58 & 9.85 \\
IAGNet & 16.16 & 19.07 & 23.92 & 14.36 & 16.90 & 21.73 \\
LASO & 22.41 & 15.90 & 30.22 & 20.06 & 8.80 & 26.84 \\
3D-ADLLM & 30.43 & 29.36 & 47.78 & 27.25 & 27.87 & 39.04 \\
\hline
VoxAfford (ours) & \textbf{34.48} & \textbf{46.15} & \textbf{61.67} & \textbf{31.87} & \textbf{49.41} & \textbf{62.18} \\
\hline
\end{tabular}
\end{table}

\begin{table}[t]
\caption{Zero-shot open-vocabulary detection results over all instances.}
\label{tab:main_instance}
\centering
\small
\setlength{\tabcolsep}{3pt}
\begin{tabular}{l|ccccc}
\hline
\multirow{2}{*}{Method} & \multicolumn{5}{c}{Full-view} \\
 & mIoU$^i$ & mAcc$^i$ & mPrec$^i$ & mRec$^i$ & mAP$_{50}^i$ \\
\hline
OpenAD-PointNet++ & 3.46 & 74.59 &11.84  & 5.84 & 0.02 \\
OpenAD-DGCNN & 3.79 & 74.42 &11.13  & 6.67 & 0.04 \\
LASO & 20.47 & 71.47 & 37.95 & 34.93 & 2.42 \\
3D-ADLLM & 30.28 & 70.66 & 40.89 & 55.93 & 27.80 \\
\hline
VoxAfford (ours) & \textbf{39.12} & \textbf{74.87} & \textbf{48.36} & \textbf{69.38} & \textbf{38.26} \\
\hline
\hline
\multirow{2}{*}{Method} & \multicolumn{5}{c}{Partial-view} \\
 & mIoU$^i$ & mAcc$^i$ & mPrec$^i$ & mRec$^i$ & mAP$_{50}^i$ \\
\hline
OpenAD-PointNet++ & 2.17 & 71.97 & 5.64 & 3.74 & 0.02 \\
OpenAD-DGCNN & 2.08 & 72.00 & 6.65 & 3.40 & 0.02 \\
LASO & 11.46 & \textbf{72.14} & 32.70 &  16.49 & 0.70 \\
3D-ADLLM & 28.72 & 68.28 &  41.71 & 47.73 & 25.63 \\
\hline
VoxAfford (ours) & \textbf{38.00} & 71.39 & \textbf{45.84} & \textbf{66.42} & \textbf{37.57} \\
\hline
\end{tabular}
\end{table}

We further evaluate on the AffordPose dataset~\cite{jian2023affordpose}, where the affordance-object combinations are entirely absent from the training set. As shown in Table~\ref{tab:ood}, VoxAfford outperforms all baselines across all metrics, improving over 3D-ADLLM by 4.81 on mIoU$^i$ and 9.08 on mAP$_{50}^i$. Since the model has never encountered these affordance-object pairings during training, the results suggest that the cross-attention fusion learns transferable alignment between affordance semantics and geometric patterns rather than memorizing specific associations from the training set.

\begin{table}[t]
\caption{Zero-shot open-vocabulary detection results on the AffordPose out-of-distribution dataset over all instances.}
\label{tab:ood}
\centering
\small
\setlength{\tabcolsep}{3pt}
\begin{tabular}{l|ccccc}
\hline
Method & mIoU$^i$ & mAcc$^i$ & mPrec$^i$ & mRec$^i$ & mAP$_{50}^i$ \\
\hline
OpenAD-PointNet++ & 7.61 & 21.36 & 64.13 & 13.01 & 0.37 \\
OpenAD-DGCNN & 8.02 & 15.83 & 66.76 & 13.52 & 0.39 \\
LASO & 34.49 & 55.44 & 76.12 & 37.88 & 8.40 \\
3D-ADLLM & 36.38 & 56.46 & 72.74 & 46.92 & 37.33 \\
\hline
VoxAfford (ours) & \textbf{41.19} & \textbf{58.57} & \textbf{77.28} & \textbf{48.57} & \textbf{46.41} \\
\hline
\end{tabular}
\end{table}

\subsection{Ablation Study}
\label{sec:ablation}

We conduct ablation experiments on the open-set zero-shot benchmark under full-view, instance-level evaluation. All variants use the same training procedure and hyperparameters unless otherwise specified.

\paragraph{Ablation on input modality and LLM backbone.}
Table~\ref{tab:ablation_input} examines whether enriching the LLM input with voxel features can resolve the spatial deficiency in the output tokens. Replacing the point cloud input with voxel features reduces mIoU$^i$ to 29.19. Concatenating both modalities improves over the voxel-only variant but still falls below our method by 7.12 on mIoU$^i$. In both cases, the geometric information must traverse the autoregressive generation process before reaching the output tokens. The sequential token-by-token generation does not preserve spatial neighborhood relations, so the output tokens remain spatially impoverished regardless of the input quality. These results validate the necessity of injecting geometric features after generation rather than before it. We also replace Phi-4~\cite{abdin2024phi} with Qwen3~\cite{yang2025qwen3} as the LLM backbone. The Qwen3 variant achieves 33.68 mIoU$^i$, surpassing the 3D-ADLLM baseline. This result confirms that the voxel-token fusion module is not specific to a particular LLM architecture.

\begin{table}[t]
\caption{Ablation on input modality and LLM backbone.}
\label{tab:ablation_input}
\centering
\small
\setlength{\tabcolsep}{3pt}
\begin{tabular}{ll|ccc}
\hline
Input & LLM & mIoU$^i$ & mAcc$^i$ & mAP$_{50}^i$ \\
\hline
Voxel & Phi-4 & 29.19 & 66.44 & 23.34 \\
Point cloud + Voxel & Phi-4 & 32.00 & 72.49 & 29.74 \\
Point cloud & Qwen3 & 33.68 & 71.41 & 31.71 \\
Point cloud & Phi-4 (ours) & \textbf{39.12} & \textbf{74.87} & \textbf{38.26} \\
\hline
\end{tabular}
\end{table}

\paragraph{Ablation on module design.}
Table~\ref{tab:ablation_component} isolates the contribution of each component. Removing the voxel-token fusion module causes the largest performance drop, reducing mIoU$^i$ to 29.64. This variant performs comparably to 3D-ADLLM, which also relies solely on LLM output tokens for mask prediction. The result confirms that the LLM output tokens alone lack sufficient spatial information for affordance localization. Removing the feature injection pathway reduces mIoU$^i$ by 5.83. Without per-point semantic conditioning from the original LLM tokens, the decoder relies entirely on the affordance prompt for guidance and loses fine-grained affordance discrimination at individual points. Replacing attention-based aggregation with concatenation lowers mIoU$^i$ by 3.15 while keeping nearly identical mAcc$^i$. Concatenation assigns equal weight to all voxel-enhanced tokens, whereas attention-based aggregation allows the model to emphasize the most relevant spatial scale for each affordance query. The difference therefore manifests primarily in spatial recall rather than per-point classification. Removing the compatibility gate decreases both mIoU$^i$ and mAcc$^i$. Without selective gating, irrelevant geometric patterns from the voxel features are injected into the token representations, which introduces noise that degrades both localization and classification.

\begin{table}[t]
\caption{Component ablation.}
\label{tab:ablation_component}
\centering
\small
\setlength{\tabcolsep}{3pt}
\begin{tabular}{l|ccc}
\hline
Variant & mIoU$^i$ & mAcc$^i$ & mAP$_{50}^i$ \\
\hline

w/o voxel-token fusion & 29.64 & 74.83 & 27.00 \\
w/o feature injection & 33.29 & 73.87 & 30.41 \\
w/o token aggregation (concat) & 35.97 & 76.54 & 33.82 \\
w/o compatibility gate (direct add) & 35.32 & 71.71 & 32.05 \\
VoxAfford (full) & \textbf{39.12} & \textbf{74.87} & \textbf{38.26} \\
\hline
\end{tabular}
\end{table}

\paragraph{Token assignment in dual-pathway propagation.}
Table~\ref{tab:ablation_assignment} compares four token assignment strategies in the dual-pathway propagation. In our default design, the prompt pathway uses voxel-enhanced tokens $\hat{\mathbf{T}}$ and the injection pathway uses original tokens $\mathbf{T}$. Reversing this assignment reduces mIoU$^i$ by 4.46. The original tokens lack multi-scale geometric structure, so they cannot generate a spatially-aware prompt for localization. Meanwhile, injecting voxel-enhanced tokens into per-point features introduces geometric signals that conflict with the point backbone's own spatial encoding. Using voxel-enhanced tokens for both pathways reduces mIoU$^i$ to 32.90. The overlapping geometric information between the prompt and the injection creates redundancy rather than complementarity. Using original tokens for both pathways yields the lowest mIoU$^i$ at 30.12, which is comparable to the variant without voxel-token fusion. Without voxel-enhanced tokens in either pathway, the spatial information from the fusion module does not reach the decoder effectively. The default assignment achieves the best balance by routing each token type to the role that matches its representational strength. The voxel-enhanced tokens carry multi-scale spatial structure suited for global prompt generation, while the original tokens preserve clean affordance semantics suited for per-point conditioning.

\begin{table}[t]
\caption{Ablation on token assignment in the dual-pathway propagation.}
\label{tab:ablation_assignment}
\centering
\small
\begin{tabular}{ll|ccc}
\hline
Prompt & Injection & mIoU$^i$ & mAcc$^i$ & mAP$_{50}^i$ \\
\hline

$\mathbf{T}$ & $\hat{\mathbf{T}}$ & 34.66 & 70.38 & 31.75 \\
$\hat{\mathbf{T}}$ & $\hat{\mathbf{T}}$ & 32.90 & 68.85 & 28.94 \\
$\mathbf{T}$ & $\mathbf{T}$ & 30.12 & 74.17 & 26.43 \\
$\hat{\mathbf{T}}$ & $\mathbf{T}$ (ours) & \textbf{39.12} & \textbf{74.87} & \textbf{38.26} \\
\hline
\end{tabular}
\end{table}

\paragraph{Multi-scale diversity in voxel-token fusion.}
Table~\ref{tab:ablation_scale} examines whether the performance gain comes from multi-scale diversity or from additional token capacity. The same-scale variant pairs all three tokens with $32^3$ features, matching the full model in token count and computational cost. This variant achieves 33.00 mIoU$^i$, which is lower than even the single $64^3$ variant at 34.82. Adding tokens without scale diversity does not improve performance and may introduce redundant representations that hinder optimization. The full multi-scale model outperforms the best single-scale variant by 4.30 on mIoU$^i$, confirming that the gain stems from combining complementary spatial granularities. Among single scales, the $64^3$ features yield the highest mIoU$^i$ because fine-grained geometric detail is most directly useful for localizing affordance boundaries. The $16^3$ features outperform $32^3$, suggesting that global shape context provides stronger affordance cues than intermediate-resolution structure. The multi-scale combination succeeds because different affordance types rely on geometric information at different granularities. The cross-attention mechanism in the fusion module allows each token to selectively retrieve the geometric patterns most relevant to its paired resolution.

\begin{table}[t]
\caption{Ablation on scale combinations and token count.}
\label{tab:ablation_scale}
\centering
\small
\begin{tabular}{lc|ccc}
\hline
Scale combination & Tokens & mIoU$^i$ & mAcc$^i$ & mAP$_{50}^i$ \\
\hline
$16^3$ only & 1 & 31.80 & 69.20 & 29.10 \\
$32^3$ only & 1 & 30.71 & 73.00 & 28.49 \\
$64^3$ only & 1 & 34.82 & 71.30 & 30.60 \\
$3 \times 32^3$ (same-scale) & 3 & 33.00 & 68.85 & 30.70 \\
$16^3$+$32^3$+$64^3$ (ours) & 3 & \textbf{39.12} & \textbf{74.87} & \textbf{38.26} \\
\hline
\end{tabular}
\end{table}

\subsection{Qualitative Results}


Figure~\ref{fig3} presents qualitative comparisons on open-set zero-shot affordance detection. As illustrated, VoxAfford consistently produces more spatially precise affordance masks than the baseline methods, closely aligning with the ground truth (GT). Methods like OpenAD and LASO frequently under-predict, yielding fragmented or incomplete regions. In contrast, our method accurately delineates the exact interaction boundaries across diverse geometric structures, demonstrating robust and tight localization capabilities without being distracted by irrelevant object parts.

\begin{figure*}
    \centering
    \includegraphics[width=0.93\textwidth]{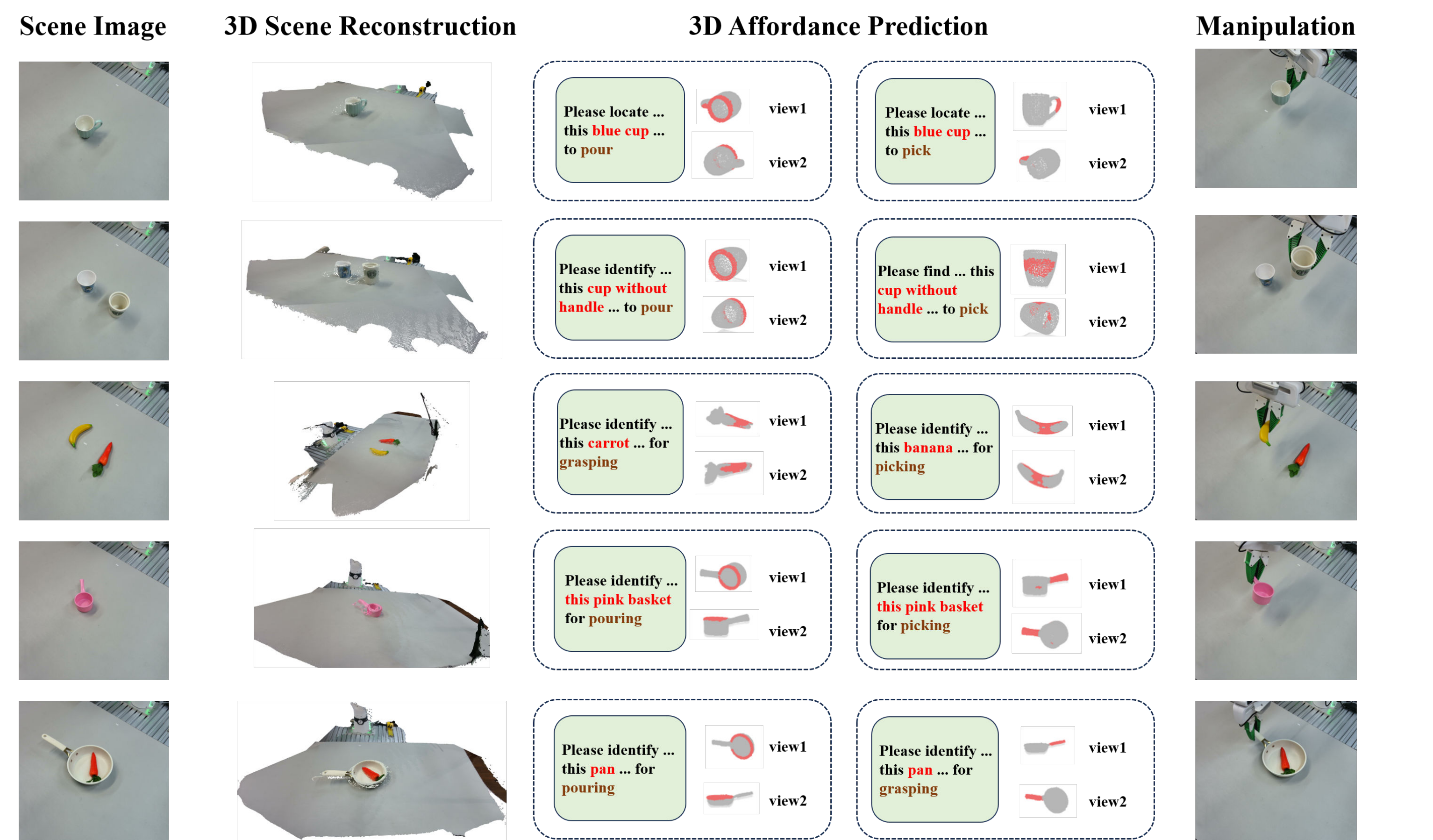}
    \caption{Real world tasks. The experimental task setup evaluated within a physical world context encompasses five unique environments. Each specific environment features entirely different objects to manipulate. The sequentially displayed columns report the initial scene images, the 3D reconstructed point clouds, the 3D affordance prediction estimates, and the corresponding manipulation execution results.
    }
    \label{fig4}
\end{figure*}

\subsection{Real Robot Experiments}
We deploy VoxAfford on a Franka Research 3 robot arm to evaluate zero-shot transfer to real-world manipulation. Three Orbbec Femto Bolt depth cameras are mounted at different viewpoints around the workspace. The three camera views are fused to reconstruct a full scene point cloud, from which the target object is segmented for affordance prediction. All test objects are novel and absent from the training set. We design five scenes with distinct objects as shown in Figure~\ref{fig4}. Each scene presents a tabletop environment with one target object. The robot receives a language query specifying the target object and the desired affordance, predicts the affordance region on the reconstructed point cloud, and executes the corresponding manipulation. Each task is repeated 10 times. Table~\ref{tab:robot} reports the success rate per object.

\begin{table}[t]
\caption{Real robot success rates on five novel objects.}
\label{tab:robot}
\centering
\small
\begin{tabular}{l|ccc}
\hline
Object & Cup w/ handle & Cup w/o handle & Fruits \\
\hline
Success rate & 8/10 & 5/10 & 6/10 \\
\hline
\multicolumn{4}{c}{\vspace{2pt}} \\
\hline
Object & Ladle & Pan & Average \\
\hline
Success rate & 5/10 & 9/10 & 33/50\\
\hline
\end{tabular}
\end{table}

VoxAfford achieves an average success rate of 66\% across all five objects. The pan and the cup with handle achieve the highest success rates because their affordance regions correspond to geometrically salient structures such as handles and rims that the voxel features can reliably capture. The cup without handle and the ladle are more challenging because their affordance regions lack distinctive geometric boundaries, requiring the model to rely more on global shape context from the coarser voxel scales for spatial discrimination. Fruits achieve moderate success. Their graspable regions are geometrically simple but vary in shape across instances, which tests the generalization of the learned affordance-geometry alignment. These results confirm that VoxAfford transfers to real-world manipulation on novel objects without additional fine-tuning.

\section{Conclusion}

We present VoxAfford, a method that injects hierarchical voxel features into LLM output tokens after autoregressive generation for open-vocabulary 3D affordance detection. The output tokens produced by existing MLLM-based methods encode affordance semantics but lack explicit 3D spatial structure for guiding mask prediction. VoxAfford addresses this limitation by pairing each output token with a designated voxel resolution and fusing geometric information through cross-attention with a learned compatibility gate. The voxel-enhanced tokens are then aggregated into a spatially-aware affordance prompt and propagated alongside per-point features for mask generation. Experiments on open-vocabulary affordance detection tasks show that VoxAfford achieves state-of-the-art performance with approximately an 8\% improvement in mIoU, and the model successfully transfers to real robot experiments.

\textbf{Limitations.}
Our method focuses on single isolated objects. Extending VoxAfford to multi-object scenes that require reasoning about inter-object spatial relations and affordance disambiguation remains an open direction for future work.

\addtolength{\textheight}{0cm}   






\bibliographystyle{IEEEtran}
\bibliography{software}

\end{document}